\documentclass[runningheads]{llncs}

\usepackage{graphicx}

\usepackage{amsmath,amssymb} 
\usepackage{color}
\usepackage{colortbl}
 \newcommand{\chapternote}[1]{{%
  \let\thempfn\relax
  \footnotetext[0]{\text{#1}}
}}

\usepackage{hyperref}

\begin{document}
\title{HANDS18: Methods, Techniques and Applications for Hand Observation}

\titlerunning{HANDS 2018 Discussion}
%
\author{
Iason Oikonomidis \inst{3}\orcidID{0000-0002-9503-3723} \and
Guillermo Garcia-Hernando \inst{1}\orcidID{0000-0003-3215-7857} \and
Angela Yao \inst{5}\orcidID{0000-0001-7418-6141} \and
Antonis Argyros \inst{2,3}\orcidID{0000-0001-8230-3192} \and
Vincent Lepetit \inst{4}\orcidID{0000-0001-9985-4433} \and
Tae-Kyun Kim \inst{1}\orcidID{0000-0002-7587-6053}
}
%
\authorrunning{I. Oikonomidis et al.}
%

\institute{Imperial College London \and
University of Crete \and
Foundation for Research and Technology \and
University of Bordeaux \and
National University of Singapore}

\maketitle              
\begin{abstract}

This report outlines the proceedings of the Fourth International Workshop on Observing and Understanding Hands in Action (HANDS 2018). The fourth instantiation of this workshop attracted significant interest from both academia and the industry. The program of the workshop included regular papers that are published as the workshop's proceedings, extended abstracts, invited posters, and invited talks.
Topics of the submitted works and invited talks and posters included novel methods for hand pose estimation from RGB, depth, or skeletal data, datasets for special cases and real-world applications, and techniques for hand motion re-targeting and hand gesture recognition.
The invited speakers are leaders in their respective areas of specialization, coming from both industry and academia.
The main conclusions that can be drawn are the turn of the community towards RGB data and the maturation of some methods and techniques, which in turn has led to increasing interest for real-world applications.


\end{abstract}

\section{Introduction}

The Fourth International Workshop on Observing and Understanding Hands in Action\chapternote{The workshop website can be found at \url{https://sites.google.com/view/hands2018} } was held in conjunction with the European Conference on Computer Vision 2018 (ECCV'18) on the 9th of September, 2018.
It was held in the main building of the Technical University of Munich (TUM) in Arcisstra{\ss}e 21, and specifically in the lecture hall Theresianum 606.
The program of the workshop included five invited talks, six full papers, three extended abstracts, six invited posters, and a short award ceremony.
This report presents in detail the proceedings of the workshop, and was not peer-reviewed.

\begin{figure}[t]
\centering
\includegraphics[width=.49\textwidth]{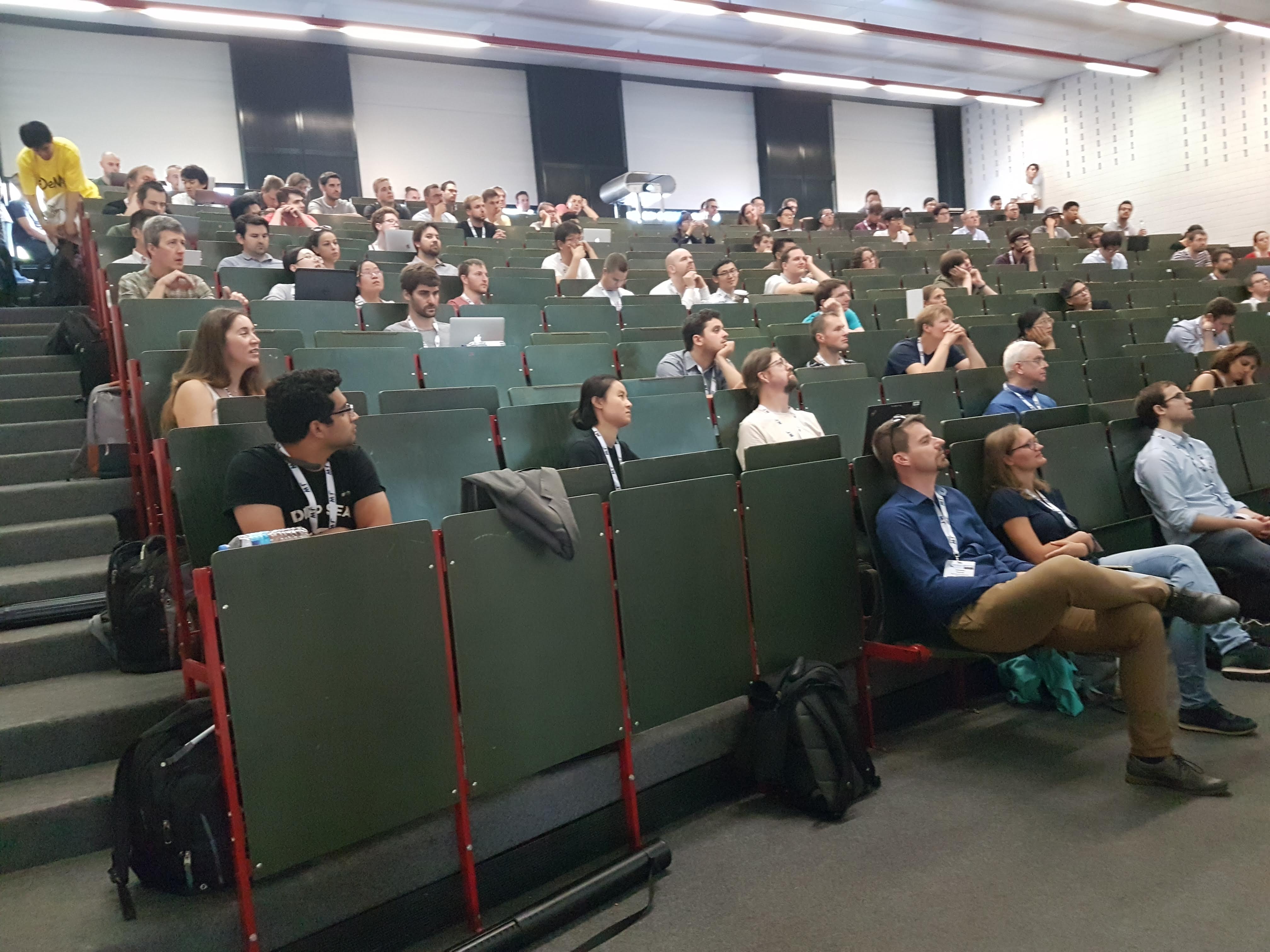}
\includegraphics[width=.49\textwidth]{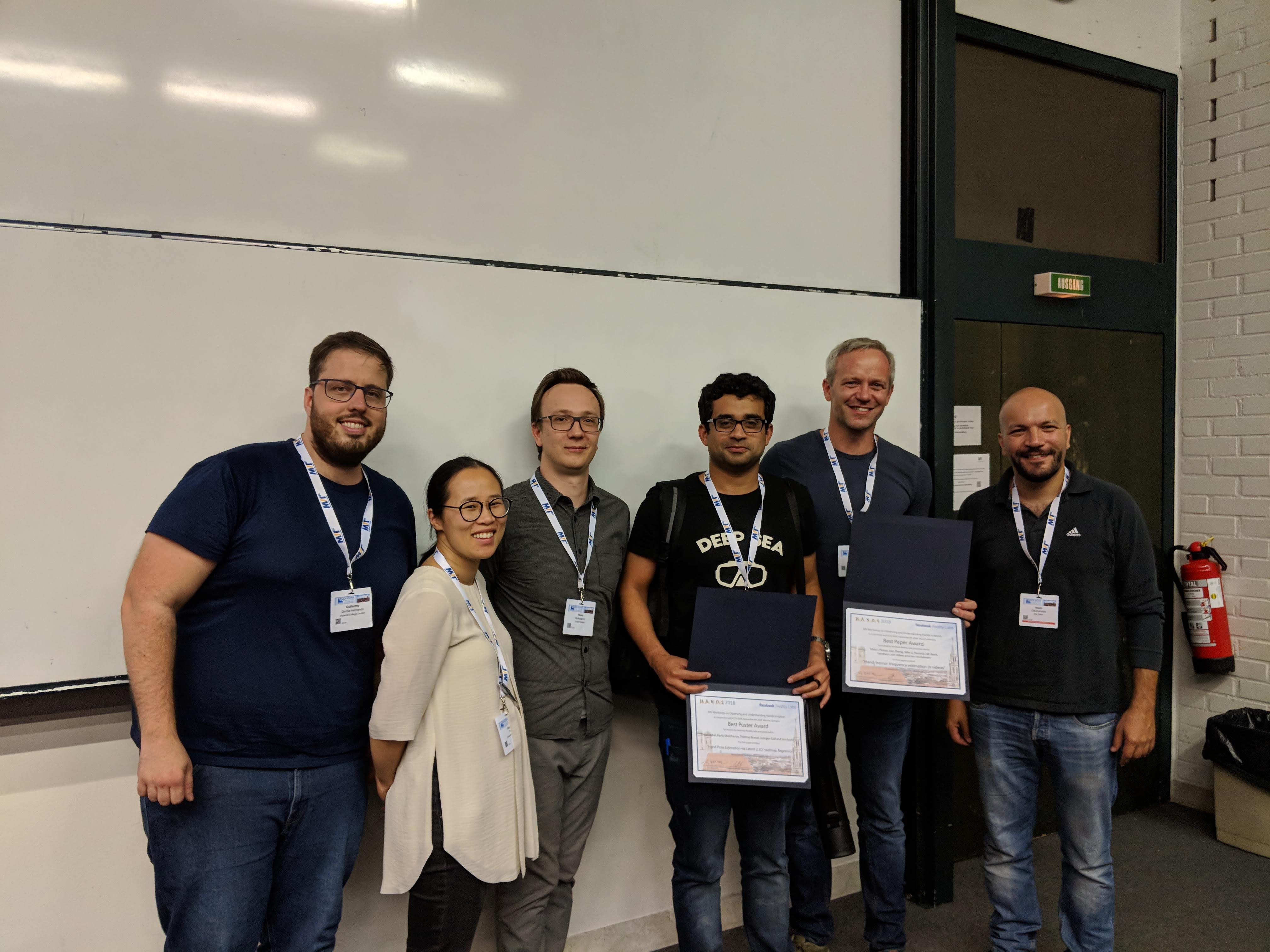}
\caption{Left: attending the talk of Professor Tamim Asfour. Right: awards photo.
From left to right: Guillermo Garcia-Hernando, Angela Yao, Pavlo Molchanov, Umar Iqbal, Jan van Gemert, and Iason Oikonomidis.
\label{pictures}}
\end{figure}

\section{Invited Talks}
There were five invited talks in the workshop, the slides of which can be found on the workshop website. 
\begin{itemize}
    \item Christian Theobalt (Max Planck Institute for Informatics) presented an overview of his recent works on hand pose estimation. He presented several different scenarios: varying input modalities including depth and regular RGB, egocentric viewpoints, and hand-object interactions  
    \cite{sridhar2014real,sridhar2016real,mueller2017real,Mueller2018}.
    \item Tamim Asfour (Karlsruhe Institute of Technology) spoke about the connection between stable grasps and humanoid locomotion states. After presenting an overview of his work on humanoid robotics, he presented the KIT whole-body human motion database \cite{mandery2015a} that can be used to efficiently solve humanoid locomotion problems.
    \item Andrew Fitzgibbon (Microsoft) talked about various optimizations over his line of work to achieve accurate real-time hand tracking performance. Among them, he focused on the Iterative Closest Point (ICP) algorithm and how it can be improved by, counter-intuitively, formulating a much larger optimization problem that includes both the model parameters and the correspondences in the same level of optimization \cite{taylor2016efficient}. The key insight is that each optimization iteration of the larger problem can take a bigger and more accurate step towards the optimum compared to standard ICP.
    \item Robert Wang (Facebook Reality Labs) talked about the process of acquiring ground truth data for hand pose estimation with markerless motion capture. After a discussion on the limitations of current approaches, he gave an overview of recent work at Facebook Reality Labs that aims to extract very accurate ground truth annotations using off-the-shelf equipment \cite{Han2018}.
    \item Andrea Tagliasacchi (Google) gave an overview of his recent hand pose estimation works, including an approach for real-time hand tracking \cite{tagliasacchi2015robust}, a method to better model the shape of the hand \cite{Tkach2016}, and a method to quickly and robustly personalized the hand model to the observed user \cite{Tkach2017}.
\end{itemize}

\section{Presented Works}

There were three types of contributions accepted for presentation in the workshop. Specifically, there were regular contributions in the form of full papers (Accepted Papers, AP), extended abstracts (EA), and invited posters (IP). Regular papers were accepted based on a peer review process. Extended abstracts were evaluated and accepted by the organizers, and had a limit of three pages. Finally,
works from the main ECCV'18 conference related to the aims and scope of the workshop were invited to be presented in the poster session of the workshop.

\subsection{Accepted Papers}

\label{aps}

The regular program of the workshop invited high quality original papers on the relevant areas.
In total, there were seven submissions that were peer-reviewed by seventeen invited reviewers. Each paper was assigned three reviewers, aiming for at least two reviews per work. Through this review process, six of the seven submitted papers were accepted for publication in the workshop proceedings. In order of sumission, the six Accepted Papers (AP) are:

\subsubsection{AP1: Hand-tremor Frequency Estimation in Videos}\cite{ID1}. This paper deals with the problem of estimating the frequency of hand tremors in patients suffering from sensorimotor disorders such as Parkinson's disease. The authors used the highly successful 2D human keypoint estimation Pose Machine method by Wei et al.~\cite{Wei2016} 
to estimate 2D wrist positions over a sequence of frames. Using these positions, two alternative approaches are proposed for the estimation of the tremor frequency. The first was named the Lagrangian approach, in which a smooth trajectory was estimated from the sequence of 2D locations. Deviations of the hand from this smooth trajectory was then used to estimate the tremor frequency. For the second Eulerian method, again the same smoothed trajectory was used, but for this approach new image features are computed around the trajectory. An analysis of these new features yields the final frequency estimation. The two proposed methods were assessed on a new collected hand tremor dataset, TIM-Tremor, containing both static and dynamic tasks. The dataset contains data from 55 tremor patient recordings including accelerometer measurements that serve as ground truth, RGB images, and aligned depth data.

\subsubsection{AP2: DrawInAir: A Lightweight Gestural Interface Based on Fingertip Regression} \cite{ID2}. ``DrawInAir'' proposes an in-air gestural recognition framework for Augmented Reality applications. The aim of this work is to enable real-time gesture recognition on lightweight devices such as a smartphones or other computationally constrained devices. 
The two main components of the framework are a fingertip localization module and a classification module that uses as input fingertip detection on subsequent frames and detects gestures. The first module is built using a fully convolutional network that outputs a heatmap of the fingertip. In contrast to common practice, an extra layer is used after the heatmap generation, applying a differentiable spatial-to-numerical transform (DSNT) \cite{Nibali2018} to convert the heatmap to numerical coordinates using a soft-argmax operation. For gesture clasiffication, a Bi-LSTM \cite{Graves2005} approach is adopted; experimental evaluation shows that this performs better than standard LSTMs \cite{Hochreiter1997}. To experimentally evaluate the proposed method, the authors collect a new dataset called ``EgoGestAR''.

\subsubsection{AP3: Adapting Egocentric Visual Hand Pose Estimation Towards a Robot-Controlled Exoskeleton}\cite{ID4}. This paper also deals with patients suffering from motor impairments -- here, the patient is assumed to have lost most of their motor abilities, and use an exoskeleton as a robotic grasp assistant. The aim of the system is to autonomously help the patient by estimating the hand pose and acting appropriately. Given that hand keypoint estimation methods usually assume that the hand is mostly free, with observed occlusions occurring from interaction with handled objects, the target scenario needs special treatment since the hand wears an exoskeleton. Towards this end, the authors propose a synthetic dataset that takes this fact into account, modeling the device and rendering hand poses with it. They adopt and adapt the approach of Wei et al.~\cite{Wei2016} comparing networks that are trained on data with and without the modeled device.

\subsubsection{AP4: Estimating 2D Multi-Hand Poses From Single Depth Images}\cite{ID5}. This paper treats the problem of 2D hand keypoint detection of two hands in a single depth image.
The authors use the Mask R-CNN object detector \cite{He2017} to detect and segment the hands in the input image. Since Mask R-CNN can be generalized to multiple human bodies, or multiple hands pose estimation, a direct approach would be to train this pipeline on the target keypoints. However, as the authors state in the manuscript, minimal domain knowledge for human pose estimation is exploited so Mask R-CNN does not adequately model joint relationships.
Moreover, another recent work \cite{Chen2017} points out that, using this strategy, key points might not be localized accurately in complex situations. To address this limitations, the authors propose a Pictorial Structure \cite{Andriluka} model-based framework. The authors evaluate the resulting system in two datasets that are generated from the single-hand datasets Dexter1 \cite{sridhar2013interactive} and NYU hand pose dataset \cite{tompson2014real} by concatenating randomly selected left and right hand images.

\subsubsection{AP5: Spatial-Temporal Attention Res-TCN for Skeleton-based Dynamic Hand Gesture Recognition}\cite{ID6}.
This paper presents the Spatial-Temporal Attention Residual Temporal Convolutional Network (STA-Res-TCN) is to recognize dynamic hand gestures using skeleton-based input. The framework consists of an end-to-end trainable network that exploits both spatial and temporal information on the input data and applies an attention mechanism on both input modalities. This results in a lightweight but accurate gesture recognition system and is evaluated on two publicly available datasets \cite{DeSmedt2016,DeSmedt2017}.

\subsubsection{AP6: Task-Oriented Hand Motion Retargeting for Dexterous Manipulation Imitation}\cite{ID7}.  This paper treats the problem of retargeting already captured motion of a human hand on another hand embodiment, such as a dexterous anthropomorphic robotic hand. The formulation follows a task-oriented approach, namely the successful grasp of the manipulated object and formulates an objective taking the task goal into account.  They proceed to learn a policy network using generative adversarial imitation learning. Experiments show that this approach achieves a higher success rate on the grasping task compared to a baseline that only retargets the motion using inverse kinematics.

\subsection{Extended abstracts}

\label{eas}

Apart from the regular papers, the workshop also had original contributions in the form of extended abstracts, with the goal of including high-potential but preliminary works.
These works were presented as posters in the poster session but are not published as part of the workshop proceedings.
Of the four submissions, three were chosen for acceptance by the program chairs. In order of submission, the three extended abstracts (EA) are:

\subsubsection{EA1: Model-Based Hand Pose Estimation for Generalized Hand Shape with Spatial Transformer Network.}
Recent work \cite{zhou2016model} has proposed using a hand kinematics model as a layer of a deep learning architecture, with the goal of making integrating differentiable coordinate transformations to enable end-to-end training. A limitation of that approach is the fact that the kinematics model has fixed parameters, making the resulting network specific for a single hand and not generalizing well to other hands. The authors of this work \cite{Li2018} extend the previous approach by adapting the kinematics parameters to the observed hand. A Spatial Transformer Network is also applied to the input image, which is shown by the experimental evaluation to be beneficial. 

\subsubsection{EA2: A New Dataset and Human Benchmark for Partially-Occluded Hand-Pose Recognition During Hand-Object Interactions from Monocular RGB Images.}
This work \cite{Barbu2018} proposes a dataset for pose estimation of partially occluded hands when handling an object. The authors collect a dataset that consists of a variety of images of hands grasping objects in natural settings.
A simple strategy enables the recording of both occluded and un-occluded images of the same grasps. The error of human annotation is evaluated using this dataset.

\subsubsection{EA3: 3D Hand Pose Estimation from Monocular RGB Images using Advanced Conditional GAN.}
This work \cite{Nguyen2018} presents a method to estimate the 3D position of hand keypoints using as input a monocular RGB image. The authors propose to decompose the problem in two stages: the first estimates a depth map from the input RGB image using a cycle-consistent GAN architecture \cite{Zhu2017}. The second stage employs a network based on Dense nets \cite{Huang2017} to regress the joint positions using as input the estimated depth map.

\subsection{Invited Posters}

\label{ips}

The organizers invited the following works from the main ECCV'18 conference related in aim to the workshop to be presented in the poster session: 

\begin{itemize}
	\item[IP1:] ``HandMap: Robust Hand Pose Estimation via Intermediate Dense Guidance Map Supervision'' \cite{Wu2018}.
	\item[IP2:] ``Point-to-Point Regression PointNet for 3D Hand Pose Estimation'' \cite{Ge2018}.
	\item[IP3:] ``Joint 3D tracking of a deformable object in interaction with a hand'' \cite{Tsoli2018}.
	\item[IP4:] ``Occlusion-aware Hand Pose Estimation Using Hierarchical Mixture Density Network'' \cite{Ye2018}.
	\item[IP5:] ``Hand Pose Estimation via Latent 2.5D Heatmap Regression'' \cite{Iqbal2018}.
	\item[IP6:] ``Weakly-supervised 3D Hand Pose Estimation from Monocular RGB Images'' \cite{Cai2018}.
\end{itemize}

\section{Awards}

Two works were given awards sponsored by Facebook Reality Labs. The best paper award was decided by the program chairs, while the best poster award was decided in a vote from selected workshop attendants including organizers, invited speakers and topic experts.

The best paper award was given to the work ``Hand-tremor Frequency Estimation in Videos'' by Silvia L Pintea, Jian Zheng, Xilin Li, Paulina J.M. Bank, Jacobus J. van Hilten and Jan van Gemert. Apart from a solid technical contribution, the work shows the applicability of the methods and techniques related to the scope of this workshop in the aid of real-world problems.  The best poster award was given to the work ``Hand Pose Estimation via Latent 2.5D Heatmap Regression'', by Umar Iqbal, Pavlo Molchanov, Thomas Breuel, Juergen Gall and Jan Kautz.  New works on hand pose estimation are increasingly turning again to regular RGB input. This work proposes a novel approach towards this end, and achieves state-of-the-art results.

\section{Discussion}

\definecolor{LightCyan}{rgb}{0.88,1,1}

\begin{table}[t]
\centering
\caption{Overview of the presented works. AP stands for Accepted Paper, EA for Extended Abstract, and IP for Invited Poster. Works are listed in the order they were presented in the respective sections.
\label{table}}

\begin{tabular}{c|c|c|c|c|c|c|c}
& RGB & Depth & Skeletal Data & Application & Dataset & Egocentric & Gesture \\
\hline
\rowcolor{LightCyan}
AP1  &   &   & x & x & x &   &   \\
AP2  & x &   &   & x & x & x &   \\
\rowcolor{LightCyan}
AP3  &   &   &   & x &   & x &   \\
AP4  &   & x &   &   &   &   &   \\
\rowcolor{LightCyan}
AP5  &   &   & x &   &   &   & x \\
AP6  &   &   & x & x &   &   &   \\
\hline
\rowcolor{LightCyan}
EA1  &   & x &   &   &   &   &   \\
EA2  &   &   &   &   & x &   &   \\
\rowcolor{LightCyan}
EA3  & x &   &   &   &   &   &   \\
\hline
IP1  &   & x &   &   &   &   &   \\
\rowcolor{LightCyan}
IP2  &   & x &   &   &   &   &   \\
IP3  &   & x &   & x & x &   &   \\
\rowcolor{LightCyan}
IP4  &   & x &   &   &   &   &   \\
IP5  & x &   &   &   &   &   &   \\
\rowcolor{LightCyan}
IP6  & x &   &   &   &   &   &   \\
\hline     
Total& 4 & 6 & 3 & 5 & 4 & 2 & 1 \\
\end{tabular}
\end{table}

Table \ref{table} provides an overview of the presented works: rows correspond to individual works (AP for Accepted Paper, EA for Extended Abstract, and IP for Invited Poster) and columns to work traits.
Specifically, ``RGB'' is marked in works that use regular RGB images a input, ``Depth'' for ones that use depth data, and ``Skeletal Data'' for works that use as input an existing estimation of the keypoints of interest. The trait ``Application'' is marked for works that solve real-world problems, ``Dataset'' for works that propose a new dataset, ``Egocentric'' for works that assume an egocentric observation of the hand, and ``Gesture'' for works that tackle the problem of recognizing hand gestures.

One conclusion that can be drawn is that the technical level of the related systems is reaching production-grade performance, with five of the fifteen works being applications. While depth-based works are still the norm, monocular RGB input is increasingly common. Furthermore, new datasets are being proposed for increasingly complex, real-world scenarios, and including stand-alone RGB input.
Towards this end, relevant scenarios involve occlusions due to hand-object interactions, other environmental factors, haptics, and robotic learning methods such as imitation and reinforcement learning. All of these scenarios should be oriented towards applications that require high precision estimates. Some of these points were mentioned in the invited talks of the workshop, and current trends on datasets point in these directions.

Two main categories of hand pose estimation approaches have been identified in the relevant literature.
Discriminative approaches, based on learning, directly map  observations to output hand poses. On the other hand, generative approaches, often based on computer graphics techniques, synthesize observations that are then compared to inputs. Then, through optimization, the hand pose that most accurately matches the observation is identified. These two classes of approaches have had small overlap over the last years. There is still much to be done towards the integration of these approaches, despite the long line of research on both categories, and also towards their integration. The resulting, third category should combine the advantages of both categories in so called hybrid approaches.
Some of the invited talks (Andrew Fitzgibbon, Andrea Tagliasacchi) focused on generative approaches while others (Christian Theobalt, Robert Wang) focused more on discriminative or hybrid approaches. Towards this integration, Andrea Tagliasacchi suggested the use of hand segmentation in a generative approach as a potential approach.

Generative approaches are focusing on more efficient methods for adaptive hand models, efficient model representations, and optimization strategies. A potentially useful observation is the fact that all modern generative approaches formulate and optimize differentiable objective functions. It is conceivable then that some of them could be used directly as loss functions in training neural networks. On the learning front, approaches that use semi-supervised learning or weak supervision \cite{Cai2018} can potentially play a big role in the immediate future. Also, techniques that enable end-to-end network training such as Spatial Transformer Networks \cite{Li2018}, Differentiable Spatial to Numerical Transform modules \cite{ID2,Iqbal2018} and kinematic layers \cite{zhou2016model} are evidently useful.

Developed methods and techniques are already being applied to solve real-world problems in healthcare, robotics, and AR/VR \cite{ID1,ID2,ID7}.
Other candidate application domains include: automotive environments, both regular and autonomous, for gesture-based interactions with the car. The surgery room for training, monitoring and aiding operations. Laboratory monitoring to record the interactions of hands and objects, and therefore, the experimental procedure. Overall, all aspects of human activity involving the manipulation of physical or virtual objects are potential candidates.

In connection to the integration of different approaches, challenging new datasets will prove useful towards the assessment of hybrid approaches. Furthermore, they can help highlight the strengths and weaknesses of discriminative and generative approaches. Following previous cases \cite{Yuan2018}, a goal for the next editions of this workshop is to organize a challenge towards this end.

%
%
%
\bibliographystyle{splncs04}
\bibliography{bibliography,program,talks}

\end{document}